\documentclass{article}

\usepackage{graphicx} 
\usepackage[usenames,dvipsnames]{color}
\usepackage{subfigure} 

\usepackage{natbib}

\usepackage{algorithm}
\usepackage{algorithmic}

\usepackage{amsmath}
\usepackage{booktabs}
\usepackage{multirow}
\usepackage{tabularx}
\usepackage[toc,page]{appendix}

\usepackage{hyperref}

\usepackage[accepted]{icml2010}

\begin{document} 
\icmltitlerunning{Regularizing Recurrent Networks}
\twocolumn[
\icmltitle{Regularizing Recurrent Networks---On Injected Noise and Norm-based Methods}

\icmlauthor{Saahil Ognawala}{ognawala@in.tum.de}
\icmladdress{Technische Universit\"at M\"unchen}
\icmlauthor{Justin Bayer}{bayerj@in.tum.de}
\icmladdress{Technische Universit\"at M\"unchen}

\icmlkeywords{machine learning, neural networks, time series, recurrent neural networks}

\vskip 0.3in
]

\begin{abstract}
\label{Abstract}
Advancements in parallel processing have lead to a surge in multilayer perceptrons' (MLP) applications and deep learning in the past decades. Recurrent Neural Networks (RNNs) give additional representational power to feedforward MLPs by providing a way to treat sequential data. However, RNNs are hard to train using conventional error backpropagation methods because of the difficulty in relating inputs over many time-steps. Regularization approaches from MLP sphere, like dropout and noisy weight training, have been insufficiently applied and tested on simple RNNs. Moreover, solutions have been proposed to improve convergence in RNNs but not enough to improve the long term dependency remembering capabilities thereof.

In this study, we aim to empirically evaluate the remembering and generalization ability of RNNs on polyphonic musical datasets. The models are trained with injected noise, random dropout, norm-based regularizers and their respective performances compared to well-initialized plain RNNs and advanced regularization methods like fast-dropout. We conclude with evidence that training with noise does not improve performance as conjectured by a few works in RNN optimization before ours. 
\end{abstract} 
\section{Introduction}
\label{introduction}
Recurrent Neural Networks are variations of multilayer perceptrons based function approximators, which are used to predict on time-series data. Such data may be text information in various languages, a musical sequence, a video, or a trend analysis in the financial domain. As training for MLP goes, the most popular techniques are all based on some form of backpropagation of weight gradients (\citet{rumelhart1988learning}). To train an RNN, the backpropagation of gradients is performed in time, on a time-unfolded representation of the network. 

When such a time series network is trained by traditional backpropagation on error gradients, it suffers from one of two peculiar analytical problems---exploding gradients or vanishing gradients. When the error gradients are backpropagated through what is essentially a set of identical weight vectors, the gradients may grow smaller (vanishing gradients) or larger (exploding gradients) exponentially fast, until they become insignificant for training purpose or lead to instability. Conceptually, the problem of vanishing gradients exists in any deep neural network that relies on propagating its error downwards to train the weights. This issue is particularly harmful in case of RNN because it damages the capability of a network to learn properties of the problem that are \emph{long-term dependent}. In simple terms, this means that due to its inherent nature of being time-series, a recurrent network needs to store not only the state representation of the input at time, $t$, but also of those seen at $t'<t$. This problem, in presence of vanishing gradients, becomes intractable for $t-t'$ exceeding a few dozens.

Due to the unstable behaviour of RNNs in dynamic space, they were not touched upon extensively until some sophisticated second-order optimization methods were introduced for feedforward neural networks \citep{martens2010deep}, that were extended to RNNs. Also groundbreaking have been the advances in form of structural solutions like Long Short-Term Memory (LSTM) \citep{hochreiter1997long} that established state-of-the-art results on text prediction tasks, pathological tasks and such. 

Till date, there have been no empirical studies on claims as the ones made in \citet{pascanu2012difficulty} that regularization of recurrent weights by means of restricting the growth of $\frac{\partial x_t}{\partial x_{t'}}$ will fail to prevent vanishing gradients. There have also not been evaluations on the standard regularization-for-overfitting techniques in MLP training applied to RNN for remembering long term dependencies. In this study, we aim to evaluate the effect of norm-based regularization methods, artificial noise injection and dropout in weights before propagating derivatives on the ability of the network to remember long term dependencies as well as convergence.
\section{Related Work}
\label{Related Work}
\citet{bengio2013advances} present an experimental study that discusses the latest optimization trends in RNNs, including gradient clipping, second order optimization methods like Hessian-free, leaky integration units (LSTMs are also discussed as a part of this), momentum tricks in simple gradient descent (SGD), powerful output probability models based on deterministic variations of Restricted Boltzmann Machines and using sparse gradients as a regularization trick. The evaluations presented in the paper above are on the same music datasets that we use in our study, in addition to the Penn Treebank Corpus of text data. 

\citet{maas2012recurrent} describe deep recurrent networks that consist of denoising autoencoders \citep{vincent2008extracting} at each time-step, to extract rich features out of audio signals by learning time-series representations from deliberately noise-ed input. The noise itself is not modelled by the autoencoder, which is the key idea behind learning a denoised input representation. 

RNNs are typically described as a set of three transition functions, viz.\ input-to-hidden, hidden-to-hidden and hidden-to-output. \citet{pascanu2013construct} delve into the matter of ``depth'' in RNNs by describing and evaluating the workings of an RNN when one or more of these three transitions are made deeper than a single layer. 

The study by \citet{hochreiter1997long} is a solution to the long term dependency problem in RNNs. In this, the authors propose a structural variation of a conventional RNN where, by adding additional short-term memory units that fire randomly, the long time-delay remembering capability of an RNN increases significantly. \citet{graves2013generating} extended the study of LSTMs by applying the idea to generate complex sequences of words in a text corpus, and handwriting patterns learned from real-valued positional information in calligraphy. \citet{zaremba2014recurrent} improved generalization in LSTMs by applying Dropout \citep{hinton2012improving} only to the non-recurrent connections. 

\citet{murray1994enhanced} present an analysis of noisy MLP training models, where the cost function is appended with a noise term to improve trajectory of the training curve, generalization of the network and increase fault tolerance from data. The results were shown to be particularly useful in the field of VLSI network design. 

The study of \citet{jim1996analysis} attempts to extend the noisy gradient descent model from feedforward networks to RNNs. The authors focus on convergence of RNNs, rather than the long term dependency problem. The noisy update model is applied to automata solving problems, which typically do not have pathologically long sequences that need to be remembered at arbitrary time delays. 

In an analysis by \citet{schaefer2008learning}, the authors claim that the widely discussed problem of long-term dependency identification in RNNs does not really exist. This claim is validated by working a pathological sequence task through an RNN, and demonstrating its performance on increasing time delays between the relevant input and output values. However, this study does not present results on standard audio, video or text corpus data that are used in other pertinent publications in RNN. 
\section{Formulation of RNNs}
\label{Formulation of RNNs}
\begin{figure}
	\includegraphics[width=\linewidth]{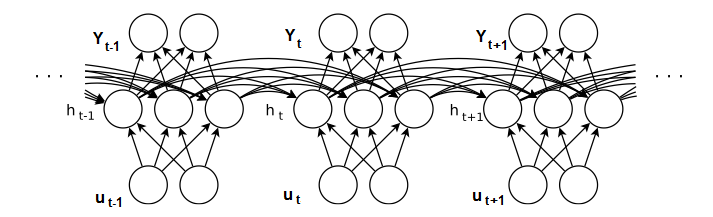}
	\caption{RNN unfolded in time. \textit{(Adapted from \citet{sutskever2013training} with permission.)}}
	\label{fig:unfolded}
\end{figure}

RNNs are semantically applicable to tasks that are based on temporal consistency. Other than universal function approximators, a way of looking at MLPs is as orthogonal representation of the input features. RNNs exploit this representation technique by duplicating hidden layers of MLP in time-steps and fully connecting the consecutive hidden layers in time. Therefore, we get an unfolded representation of RNNs in time as shown in Fig.~\ref{fig:unfolded}. 

We, hence, define an RNN as
\begin{align}
\label{eqn:Youtput}
Y_t &= \sigma_o(W_{ho}X_t + b_o) \\
\label{eqn:Xhidden}
X_t &= \sigma(W_{hh}X_{t-1} + W_{ih}u_t + b_h)
\end{align}
At time-step $t$, $u_t$ is the input, $X_t$ is the activation of the hidden layer, $h_t$, and $Y_t$ is the output of the network.
The complete parameter set of the model is given by the input-to-hidden weights, $W_{ih}$, hidden-to-hidden weights, $W_{hh}$, hidden-to-output weights, $W_{ho}$, hidden layer bias, $b_h$, and output layer bias, $b_o$. $\sigma_o$ and $\sigma$ are the non-linear activation functions at the output and hidden layers, respectively.

\section{Exploding Gradients and Effect on Long-term Dependencies}
\label{Exploding and Vanishing Gradients}
\citet{bengio1994learning} and \citet{pascanu2012difficulty} explain the dynamics of the weight training using backpropagation through time in RNNs.

Consider the error function, $\varepsilon$, applied on the outputs of RNN. Calculating error gradient
\begin{align}
\frac{\partial \varepsilon}{\partial \theta} &= \sum_{1 \leq t \leq T} \frac{\partial \varepsilon_t}{\partial \theta} \\
\label{eqn:partial_derivative}
\frac{\partial \varepsilon_t}{\partial \theta} &= \sum_{1 \leq k \leq t} \frac{\partial \varepsilon_t}{\partial X_t} \frac{\partial X_t}{\partial X_k} \frac{\partial X_k}{\partial \theta}
\end{align}
Where, $\theta$ is the concatenated matrix of $W_{ih}$, $b_h$, $W_{hh}$, $W_{ho}$ and $b_o$.

It is clear from Eq.~(\ref{eqn:partial_derivative}) that derivative of loss function at every time-step, $t$, is affected by the the activations at time-steps $k<t$.
 
Furthermore, consider the term $(\partial X_t / \partial X_k)$ on the right hand side of Eq.~(\ref{eqn:partial_derivative}) 
\begin{align}
\frac{\partial X_t}{\partial X_k} = \prod_{t \geq j \geq k+1} \frac{\partial X_j}{\partial X_{j-1}}
\end{align}
The multiplication of the real valued derivatives at time-steps $k<t$ successively for all indices $t$ in Eq.~(\ref{eqn:partial_derivative}) may lead to the norm of the product growing very large or vanishing to zero, exponentially fast in time. This is harmful as far as storing long term time dependencies goes, because by the time the error gradient at $k$ would have been propagated to $j<<k$, the norm explosion or vanishing may have made the training regime unsuitable for any meaningful updates.

This compounding of the error gradient can happen in one of two opposite directions, both depending on the largest eigenvalue (spectral-radius), $\rho$, of the recurrent weight matrix. If the spectral radius is much less than 1, the gradient might vanish over time (if using a sigmoid-like non-linearity). On the other hand, if the spectral radius is bigger than 1, the gradient might explode over time.
\section{Demonstration with a Simple Regime}
\label{Simple Demo}
Let us demonstrate the delicate nature of training a recurrent weight matrix, using an over-simplified architecture (a more expansive explanation, also from a dynamical systems perspective, can be found in \citet{pascanu2012understanding}).

In Eq.~\ref{eqn:Xhidden}, assume that there is no new input coming at every time-step, so that the second term with $u_t$ becomes unnecessary. Furthermore, assume that $X$ is a single dimension variable, which means that $W_{hh}$ and $b_h$ have dimensions [1, 1] and [1] respectively.

\begin{figure}
\includegraphics[width=\linewidth]{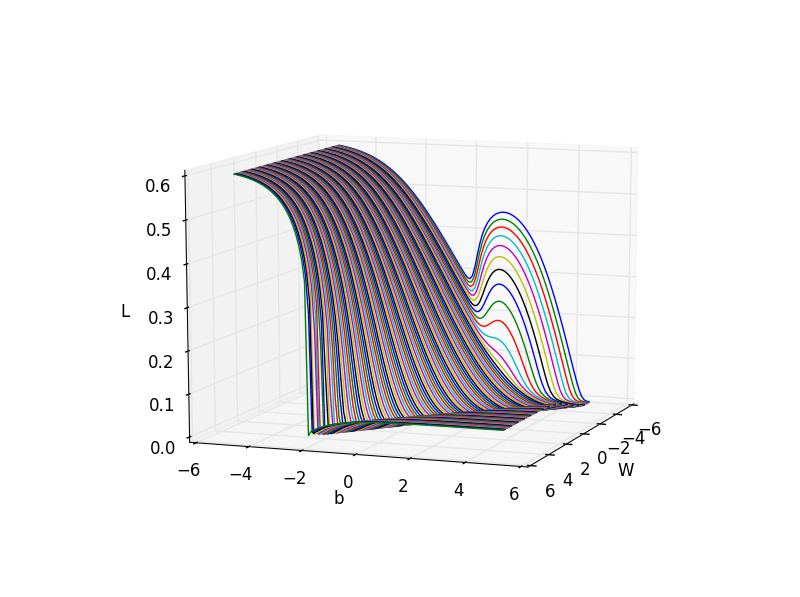}
\caption{Training regime of a simple RNN. $W$: weight, $b$: bias, $L$: squared loss}
\label{fig:simple_opt_noreg}
\end{figure}

Our objective, then, is to start $x$ with a zero value and reach a given target value, $z$, in a set number of time-steps. Fig.~\ref{fig:simple_opt_noreg} shows the training graph over 10000 different initialization sets of $W$ and $b$. On the third axis, $L$ represents the squared loss of the model.

The steep wall perpendicular to the parameter space represents an explosion in gradients of the loss function. When the largest eigenvalue in the parameter matrix explodes, the curvature of the error surface compounds too, which is what the wall illustrates.

The thing most noteworthy is that when the search routine is at a point on the top surface of the error curve, it makes its next step in a direction perpendicular to the face of the wall. Depending on the learning rate, it might then fall to ground beyond the valley where the error reaches its minimum. This is not such a big problem, because the search must come back to the valley region, given itself to explore the ground region. Note, however, that is only until the search direction collides onto the wall again, at which point a small change in the norm of the update would take the search back to the top of the hill to repeat the entire search process.

\begin{figure}
\includegraphics[width=\linewidth]{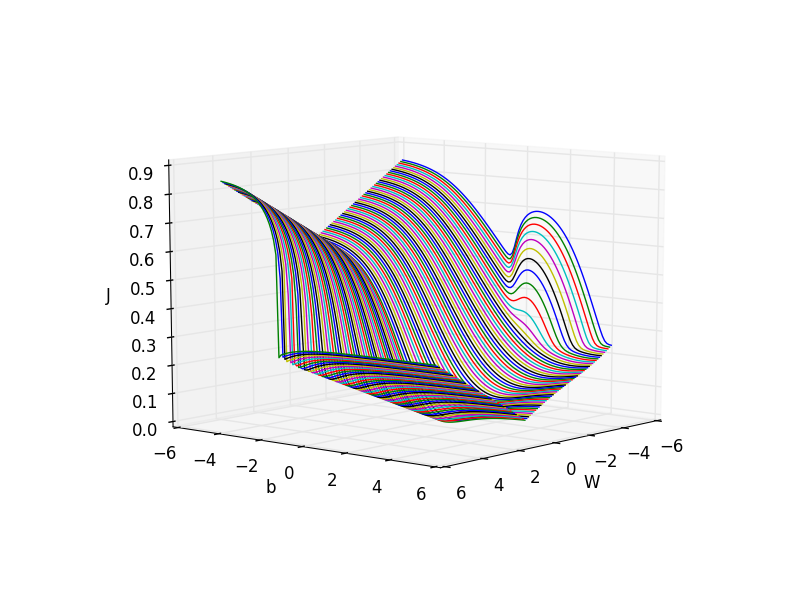}
\caption{More desirable training regime of a simple RNN.}
\label{fig:desirable_opt}
\end{figure}

The key, then, is to have a method that would smoothen the minima valley and decrease the slope of the steep wall so as to allow optimization to move in a less arbitrary fashion given a sufficiently small learning rate. A more acceptable routine may look like the one shown in Fig.~\ref{fig:desirable_opt}. 

\section{Existing Solutions}
\label{Existing Solutions}
\subsection{Initialization and Momentum Tricks}
\label{Initializing and Momentum Tricks}
Momentum (\citet{polyak1964some}) with SGD method has the added advantage of preserving the directions of consistent change over multiple updates. The persistent change in directions can be thought of as the dominant velocity in which the update moves during the optimization process. \citet{sutskever2013importance} describe Gradient descent with momentum as
\begin{align}
v_{i+1} &= \mu v_i - \epsilon \nabla \varepsilon(\theta_i) \\
\theta_{i+1} &= \theta_i + v_{i+1}
\end{align}
Where, $\theta_i$ is the weight matrix after $i$ updates, $v_i$ is the $i^{th}$ update value, $\mu$ is the momentum, $\epsilon$ is the step rate of learning and $\nabla \varepsilon(\theta_i)$ is the partial derivative of the error function w.r.t.\ the parameter $\theta_i$. 

\citet{nesterov1983method} introduced Nesterov Accelerated Gradient (NAG) method for effective velocity preservation in optimization process. In the manner of classical momentum, NAG can be formalized as
\begin{align}
v_{t+1} &= \mu v_t - \epsilon \nabla \varepsilon(\theta_t + \mu v_t) \\
\theta_{t+1} &= \theta_t + v_{t+1}
\end{align}

The small, but key, difference between classical momentum method and NAG is that in the latter, first a partial update to the parameters is done using the last update value, and then the gradient calculation is done for the next update.

The second trick presented by \citet{sutskever2013importance} is related to the random initialization of the hidden-to-hidden and input-to-hidden weight matrices. The sparse-ifying technique presented here is inspired by \citet{martens2010deep}, where all but 15 (or some $k$) connection weights are set to zero, and the rest are sampled from a Gaussian distribution. The reasoning behind this weight setting has been that a sparse connection matrix would help to diversify the incoming connection from a lower layer.

As a second initialization step, the spectral-radius is kept close to 1, so as to decrease the possibility of the gradients exploding or vanishing over a long time delay, when using sigmoid transfer function.

\subsection{Echo-State Networks}
\label{Echo-State Networks}
It has been argued by \citet{jaeger2004harnessing} that a random draw from a pre-determined distribution can be used to set the input-to-hidden and hidden-to-hidden connection weights, instead of learning them iteratively. This method, however, is not applied to the hidden-to-output layer connections, which are trained using closed form solutions that involve calculating the pseudo-inverse of a Hessian matrix.

A completely random draw without controlling the distribution parameters might be harmful for setting such weights, though. For instance, if the spectral radius of the hidden-to-hidden weight matrix is much higher or lower than 1, there is a clear possibility that the long term dependency effects are either intractable or vanish, respectively, over time. Hence, we follow the general rule that the spectral radius of the hidden-to-hidden weight matrix is restricted to be close to 1 (1.1, 0.9 etc.) and the input-to-hidden weights are drawn with a small standard deviation of about 0.001.

\subsection{Hessian Free Optimization}
\label{Hessian Free}
\citet{martens2010deep} propose a second order Hessian-Free (HF) optimization method, inspired by Newton's method, to train deep neural networks with random initializations. HF method obviates the need for pre-training in deep models, which was previously thought to be the most promising way of starting the optimization process, due to the presence of deep pathologies \citep{hinton2006fast, hinton2006reducing}.

With respect to the objective function, $f(\theta)$, HF concerns itself with optimizing a simpler sub-objective of $f(\theta)$ by finding local approximations to it. This is done as follows---for a parameter update from $\theta_n$ to $\theta_{n+1}$, it optimizes a sub-objective function
\begin{align}
q_{\theta_n}(\theta) = M_{\theta_n}(\theta) + \lambda R_{\theta_n}(\theta)
\end{align}
The term, $M_{\theta_n}(\theta)$, represents a quadratic approximation to $f(\theta_n)$. Normally, $M_{\theta_n}(\theta)$ is chosen to be the Taylor-series expansion of $f$ to second-order terms. This is the same expansion term that is used for Newton's optimization methods with the key difference that there are no additional assumptions like a low-rank matrix. This would, typically, make the optimization harder since it would involve an inversion of a large matrix. What differentiates HF from other second order optimization methods is that it is made possible to partially optimize $q_{\theta_n}(\theta)$ by conjugate gradient method, instead of gradient descent. 

The term $R_{\theta_n}(\theta)$ is a regularization function that penalizes the solution as it moves farther away from $\theta_n$ (this modification to the HF method of \citet{martens2010deep} was proposed by \citet{sutskever2013training}).

\subsection{Long Short-Term Memory (LSTM)}
\label{LSTM}
\begin{figure}
	\includegraphics[width=\linewidth]{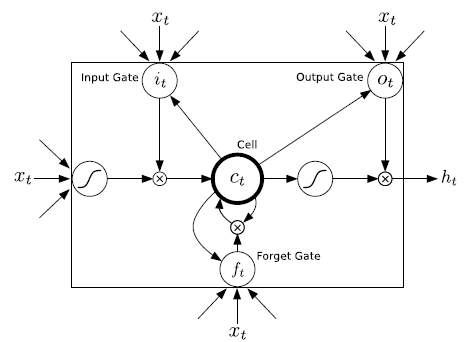}
	\caption{LSTM cell schematic \citep{graves2013generating}}
	\label{fig:lstm}
\end{figure}
While not particularly a solution to the exploding/vanishing gradients problem, LSTMs \citep{hochreiter1997long} have been systematically proven \citep{graves2013generating} to have state-of-the-art performance on sequence generation and long-range time series prediction tasks. LSTM alleviates the temporal dependency preservation problem of plain RNNs by structurally modifying the naive neural nodes of the RNN model to produce a more complex LSTM memory cell. 

LSTM cell consists of the following novel links, as in Fig.~\ref{fig:lstm}, in addition to the conventional hidden units
\begin{itemize}
	\item \emph{Input gate} to control the in-flow of an input vector into the hidden state. Takes a value from $[0, 1]^N$.
	\item \emph{Output gate} to control the out-flow of a hidden state activation to the next layer of LSTM-RNN. Takes a value from $[0, 1]^N$.
	\item \emph{Forget gate} to control the value retention of a memory cell. This link uses the input vector and hidden activation value to determine whether the activation is fed back to the unit for retention over longer time sequences. Takes a value from $[0, 1]^N$.
\end{itemize}

The original LSTM by \citet{hochreiter1997long} uses SGD for training, but it suffers from the exploding gradient problem. In order to solve that, the solution of \citet{graves2013generating} uses gradient clipping technique to limit the norm of the gradients and hence stop them from growing too large with time. Even so, the structural complexity of LSTM memory units makes it difficult to implement and harder to train on most systems that do not allow calculation of arbitrary gradients. 

\subsection{Fast-Dropout RNNs}
\label{Fast-Dropout}
\citet{wang2013fast} suggest an approximation for dropout \citep{hinton2012improving} in deep neural networks. The suggestion is to treat every neuron as a random variable, whose incoming connections are randomly set to zero, with a probability of $1-p$. It would be safe to assume that the nature of such a random variable would tend to be Gaussian over sufficiently large number (approximately 10, or more) of incoming connections. The resulting models had orders of magnitude better training times than a naive dropout approach, and the test results matched, and were sometimes better than those of plain MLPs.

\citet{bayer2013fast} verified the validity of the fast-dropout approach on RNNs. This was done by concatenating the input-to-hidden and hidden-to-hidden weights into a single array, and applying the same approximation to the incoming connections as in \citet{wang2013fast}. Fast-dropout applied to RNNs, works as a regularizer, because the Gaussian approximation of the dropout term leads to a local derivative of the random variable representation of the node, that acts as an additive regularization term. 

The results of Fast-dropout, when applied with the initialization tricks of Sec.~\ref{Initializing and Momentum Tricks} on standard music datasets, produces state-of-the-art results. 

\section{Norm-based Regularizers}
\label{Norm-based Regularizers}
The first method of regularization in RNN that we evaluate is Tikhonov regularization \citep{bishop1991improving} on input-to-hidden, recurrent and hidden-to-output weight matrices. It has been claimed in previous RNN related works \citep{pascanu2012difficulty} that L1 and L2 penalties on the weight matrix, when added to the cost function of the estimator, may work against improving the long-term dependency remembrance of the network and only partially alleviate the exploding gradients problem. 

Using the same example for demonstration as in Sec.~\ref{Simple Demo}, we illustrate the effect of L1 and L2 regularizers on the training regime of a time-series network. 

\section{Stochastic Noise Injection}
\label{Stochastic Noise Injection}
Noise injection is used as a regularization method in feedforward-only neural networks (\citet{bremermann1991brain}, \citet{flower1993summed}, \citet{jabri1992weight}) to improve generalization. The motive behind adding stochastic noise of different natures to the synaptic weights is to improve fault tolerance in the input and gracefully handle unseen data during prediction.

Adding noise to the weights during optimization works as a regularizer by, essentially, converting the state-space search into a search in a more coarse region of the weight space than what would have been without the additional noise. This property of noisy training has been exploited for training the recurrent weights in RNNs too. By adjusting the weight space to a grainier region, not only are we promised faster convergence but also a cure for the exploding gradients problem. A detailed analysis of Gaussian noise injection in recurrent weight matrix and its behaviour as a regularizer is given in appendix \ref{noise_analysis}. 

In RNNs, the work of \citet{jim1996analysis} demonstrate application of stochastic noise to the recurrent layers, much the same way as feedforward MLPs. In the following subsections, we use the additive and multiplicative noise addition model by \citet{jim1996analysis} to evaluate the performance of a recurrent network in terms of preserving long term dependencies in musical chord sequences. Our analysis of the noisy recurrent weight training model is followed by noisy input-to-hidden weight model.

\subsection{Noise in Recurrent Weights}
\label{Recurrent noise}
The first type of noise injection we analyze is in the recurrent weight matrix. In all the analyzed noisy training methods, we restrict ourselves to non-cumulative noise models. In non-cumulative noise methods, the intensity of noise injected at each time-step, $t$, is independent of the amount of noise injected at $k<t$. As we saw earlier, backpropagation-through-time in RNNs trains essentially the same set of weights in time-space and, hence, we postulate that cumulatively increasing the noise intensity in  time space might decrease the convergence performance of the network.

Other than the cumulative nature of the recurrent weight noise, there are two main considerations for deciding the nature of noise that must be injected at each recurrent layer
\begin{enumerate}
\item Should the same noise vector be inserted at every time-step in the unrolled representation of the network (per-sequence noise) or a different noise vector be sampled for every time-step (per-time-step noise)?
\item Should the noise be a multiplicative factor of the state of weight vector (multiplicative noise) or simply an additive noise vector sampled from a given distribution (additive noise)?
\end{enumerate}

\subsubsection{Additive Noise}
\label{Additive Noise}
Additive noise in recurrent weights at time-step, $t$, is given by
\begin{align}
W_{hh_{t}}^\ast \leftarrow W_{hh_{t}} + \Delta_{hh_{t}}
\end{align}
$W_{hh_{t}}^\ast$ is the modified version of $W_{hh_{t}}$ after adding the noise term. The noise vector, $\Delta_{hh}$ is chosen from a standard normal distribution
\begin{align}
\Delta_{hh} \sim \mathcal{N}(0, \sigma) \nonumber
\end{align}
In the per-time-step recurrent noise model, we sample a new noise vector, $\Delta_{hh_t}$ for every time-step in the unrolled-representation for every iteration of weight update in the optimization process. 
In the per-sequence recurrent noise model, we sample a new noise vector, $\Delta_{hh}^\ast$ for every iteration in the optimization process and add the same noise to each time-step in the network. 

\subsubsection{Multiplicative Noise}
\label{Multiplicative Noise}
Multiplicative noise in recurrent weights, analogously, is given by
\begin{align}
W_{hh_{t}}^\ast \leftarrow W_{hh_{t}} + W_{hh_{t}}\Delta_{hh_{t}}
\end{align}
The nature of $\Delta_{hh_t}$ is the same as before. 

As with additive noise, multiplicative noise is also evaluated on the two variants of per-time-step noise and per-sequence noise models. 

In both, additive and multiplicative noise models, the perturbation of the weight matrix is done only during the optimization period, and not during forward propagation. During weight training, the original values of the weight matrices are preserved even as noise is added for the gradient calculation for backpropagation-through-time.

\subsection{Noise in Feedforward layers}
\label{Feedforward noise}
As with noise in the recurrent weight matrix, we would like to close the loop on experimentation by applying the noisy weights training on the feedforward connections too.

During training of feedforward connections with backpropagation of gradients, we use the following weight formulae for noisy weights
\begin{align}
W_{ih_{t}}^\ast &\leftarrow W_{ih_{t}} + \Delta_{ih_{t}} \\
W_{ho_{t}}^\ast &\leftarrow W_{ho_{t}} + \Delta_{ho_{t}}
\end{align}

We only work with per-time-step noise model for feedforward layers. 

\section{Dropout as a Regularizer}
\label{Dropout as a Regularizer}
Random dropout in MLP connections is used as a generalization technique \citep{hinton2012improving}, that works by preventing co-adaptation of multiple features in the training set. A variation of dropout in the activation units is DropConnect \citep{wan2013regularization}, where random elements from the weight matrix are dropped instead.

We use the DropConnect model on the recurrent weight matrix to try to improve the long-term dependency preserving tendency of our network. As with stochastic noise reduction, dropout in recurrent weights can be applied in two different ways
\begin{enumerate}
	\item A possibly unique set of weights are dropped out at every time-step (per-time-step dropout).
	\item Same set of weights are dropped out at every time-step (per-sequence dropout).
\end{enumerate}
After searching over the range 0--1, we find the best dropout rate suitable for the recurrent connections.

\section{Experiments}
\label{Experiments}
\subsection{Datasets}
\label{Datasets}
For evaluating the proposed regularization techniques, we use musical datasets. These are notes based representation of score sheets from four sources---JSB Chorales (harmonized chorales of J.S.~Bach), Piano-midi.de (classical music from different sources), Nottingham (folk tunes) and MuseData (classical music).

The dimensionality at each time-step for all four datasets is 88. After dividing the original dataset into training, validation and testing sets (approximately 60\%--20\%--20\% respectively), we split the training and validation samples into chunks of 100 time-steps each. We choose this number because in our experience, for a dataset such as music scores, a length of 100 is long enough to make remembering long term dependencies a necessity while at the same time not making it unreasonably difficult for a network to do so. For samples that are smaller than 100 steps long, we pad them with zeros at the front.

We do no such splitting or prefixing for the test dataset, and use the original sized data chunks for prediction. 

\subsection{Model Description}
\label{Model Description}
Our setup for all four polyphonic music datasets consists of one hidden layer of neurons at each time-step of the RNN. The number of hidden units in the layer is enumerated in the appendix \ref{Hyper params}. The hidden units use the hyperbolic-tangent (tanh) non-linearity and the output nodes use sigmoid. The model parameters are tasked with describing the random variable, $\mathbf{y}$, such that 
\begin{align*}
y_{t, i} = p(x_{t, i} | \mathbf{x}_{1:t-1})
\end{align*}
Where $x_{t, i}$ denotes the state of note $i$ at time-step $t$ which, if present, is $1$ and $0$ otherwise.

The loss function which is optimized by this RNN is a mean cross-entropy (CE) loss over all time-steps
\begin{align*}
\varepsilon(\theta) = \frac{1}{T-1} \frac{1}{N} \sum_{i, t, k}x^{(k)}_{t, i}\log{y^{(k)}_{t-1, i}} \\ + \ (1-x^{(k)}_{t, i})\log{(1-y^{(k)}_{t-1, i})}
\end{align*} 
$i$ denotes the note index, $t$ denotes the time-step and $k$ denotes the training sample index. 

\subsection{Results}
\label{Results}
On the four datasets, we report the average CE errors in Tab.~\ref{tab:Results-table}. The results for RNN with norm-based regularizer (RNN-NBR), per-time-step noise (RNN-N), per-sequence noise (RNN-NS), multiplicative noise per-time-step (RNN-MN), multiplicative noise per-sequence (RNN-MNS), dropout per-time-step (RNN-DO), dropout per-sequence (RNN-DOS) and feedforward noise (RNN-FF) are given compared to plain RNNs (with initialization in correct regime) and fast dropout RNN (RNN-FD). Advanced training methods like fast dropout and RNN-NADE \citep{boulanger2012modeling} perform measurably better on this data.

We see that injecting stochastic noise or randomly dropping out weights in recurrent layers during training does not necessarily improve the performance of the RNN training or generalization to the test set. In fact, for most datasets, simply tuning the initialization parameters viz.~standard deviation of the weight parameter sampling, sparsification of the weight matrix and spectral radius of the recurrent weight vectors, provides better test performance on the musical datasets, than using the noise injection techniques. 
\begin{table}[h]
		\caption{Test set results on polyphonic musical datasets}
		\label{tab:Results-table}
		\begin{tabular}{|l|c|c|c|c|}
			\toprule
								  & JSBC	    & Not.       & P-midi        & Muse \\ \hline 
			Plain-RNN             & 8.58        & 3.43       & 7.58          & 6.99     \\
			RNN-FD                & 8.01        & 3.09       & 7.39          & 6.75     \\ \hline 
			RNN-NBR               & 8.83        & 3.70       & 7.78          & 8.62     \\
			RNN-N                 & 8.92        & 3.56       & 7.66          & 8.40     \\
			RNN-NS                & 8.96        & 3.58       & 7.74          & 8.40     \\
			RNN-MN                & 8.64        & 3.51       & 7.71          & 8.13     \\
			RNN-MNS               & 8.64        & 3.50       & 7.70          & 8.12     \\
			RNN-DO                & 8.48        & 3.49       & 7.65          & 7.98     \\
			RNN-DOS               & 8.55        & 3.57       & 7.67          & 8.00     \\
			RNN-FF                & 8.67        & 3.54       & 7.69          & 8.10     \\ \hline 
		\end{tabular}
		
\end{table}

\begin{figure}
	\includegraphics[width=\linewidth]{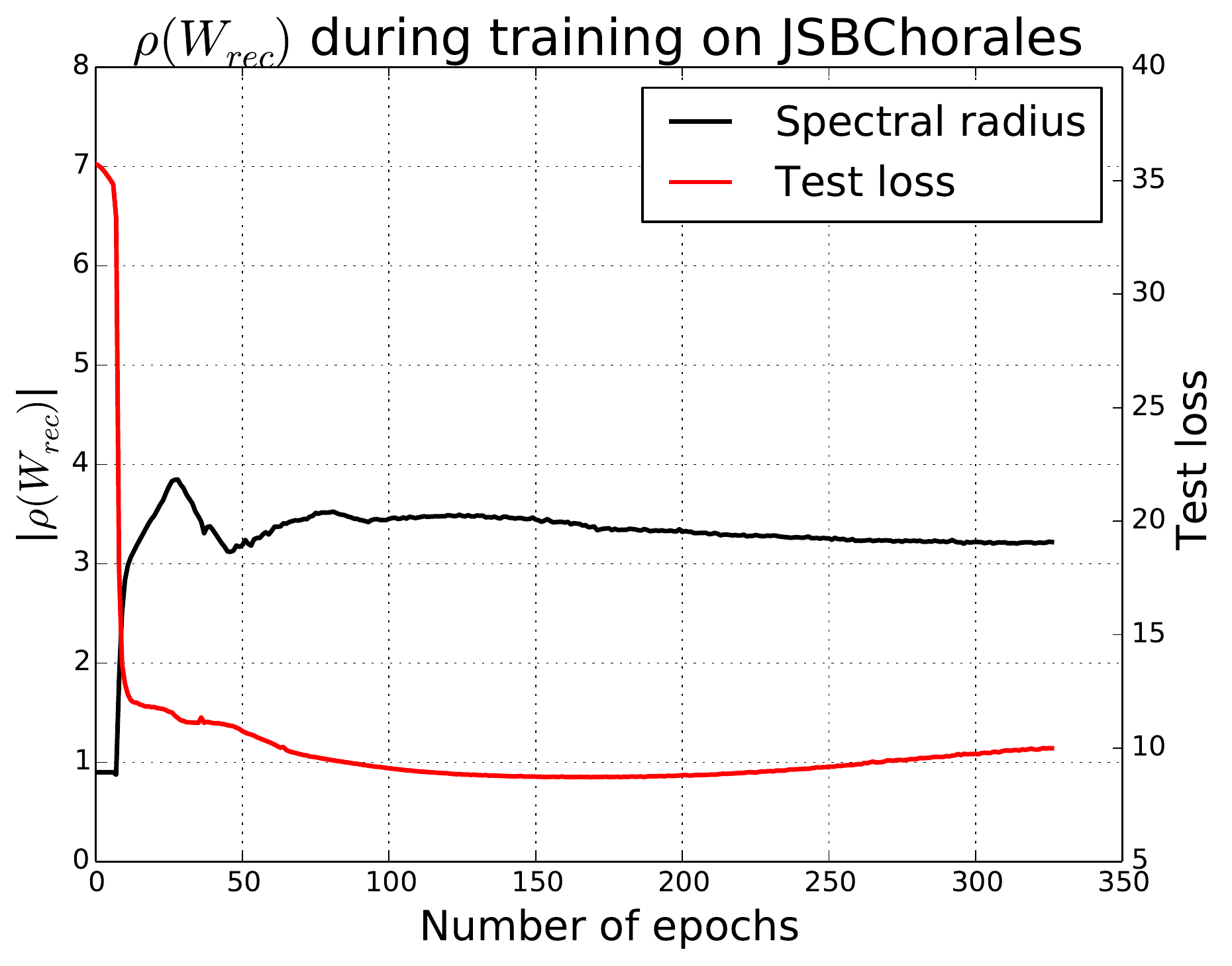}
	\caption{Training with multiplicative noise per-sequence}
	\label{fig:spectralradius_multseq}
\end{figure}

\begin{figure}
	\includegraphics[width=\linewidth]{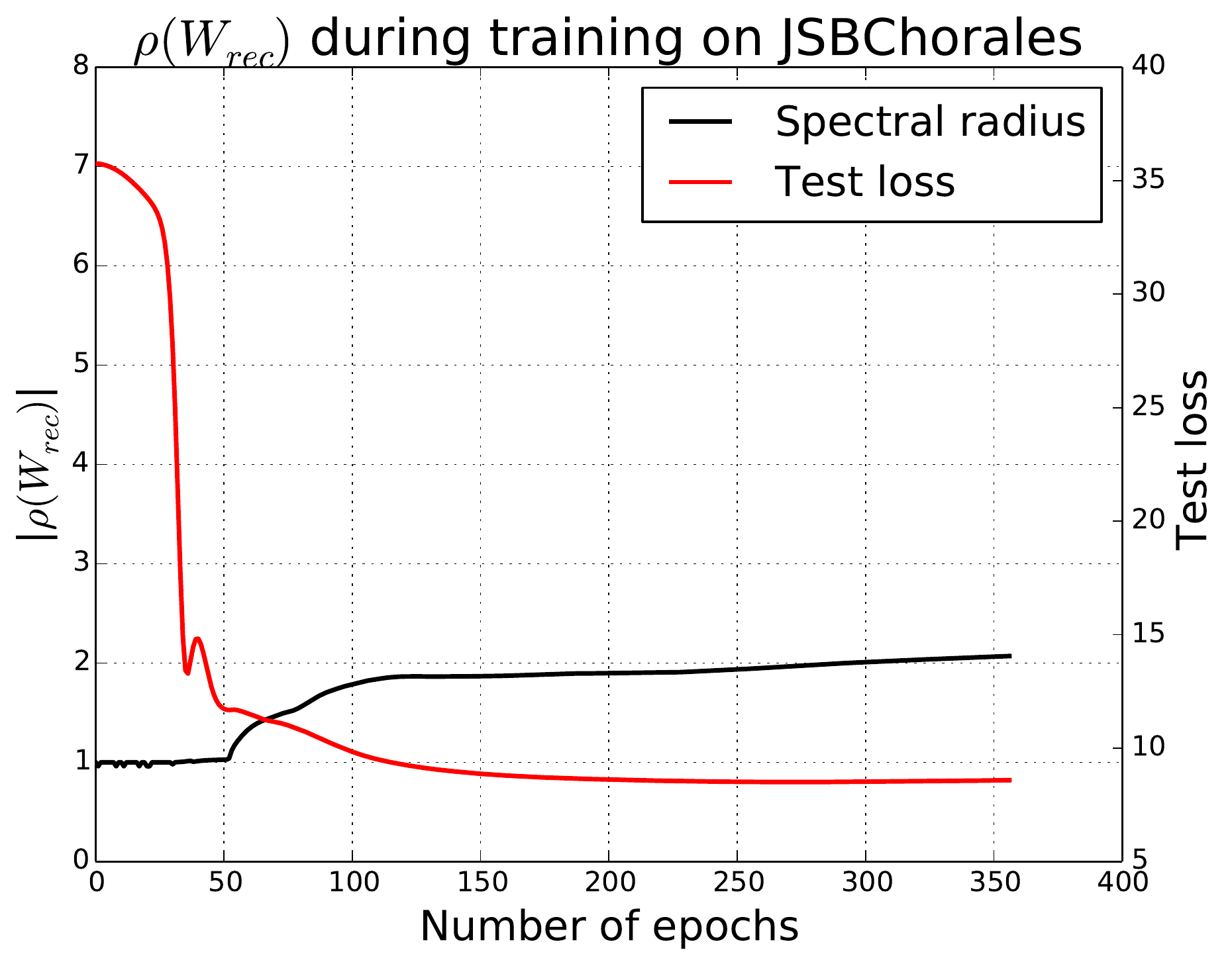}
	\caption{Training with dropout}
	\label{fig:spectralradius_dropout}
\end{figure}

\begin{figure}
	\includegraphics[width=\linewidth]{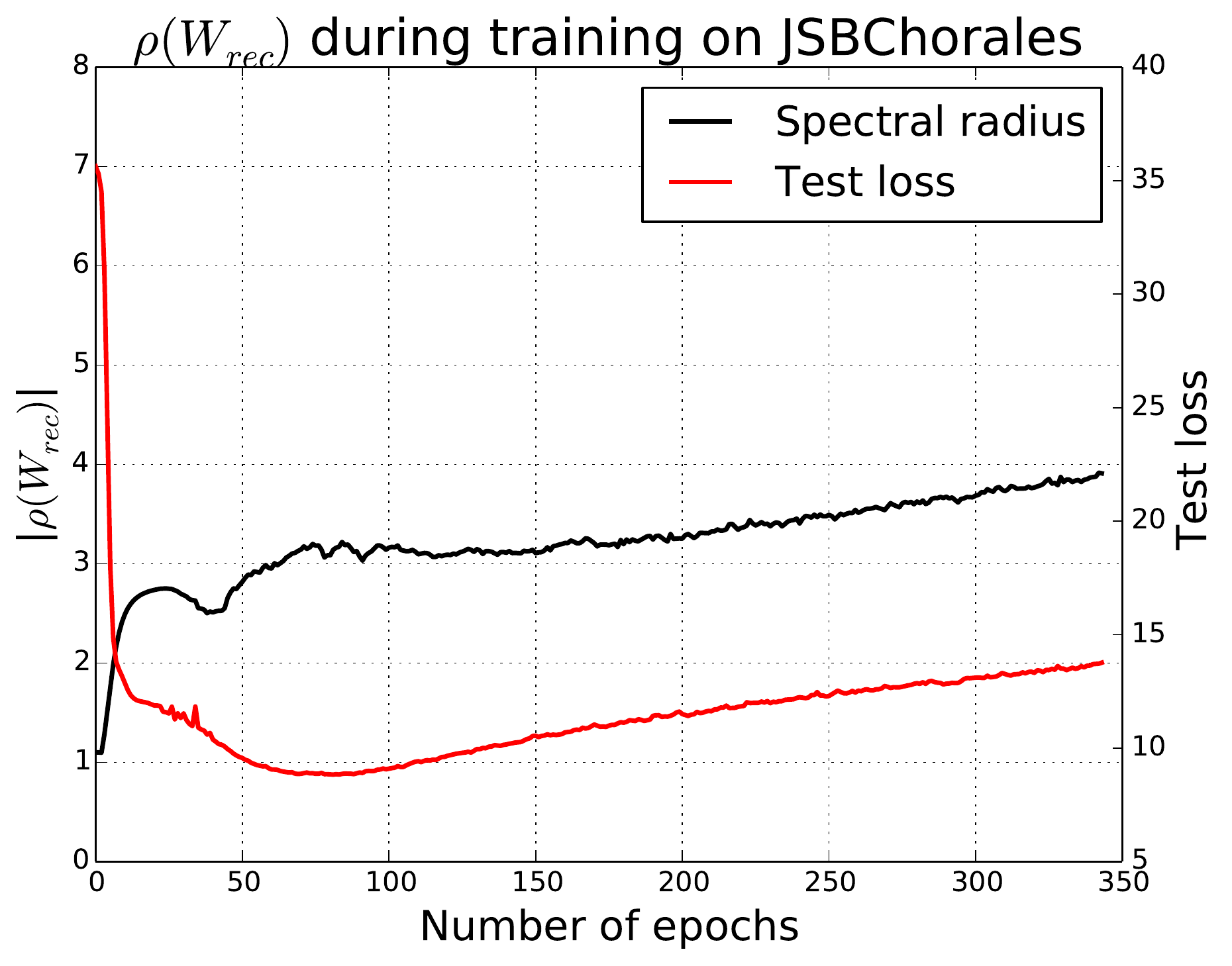}
	\caption{Training with L2 regularizer}
	\label{fig:spectralradius_reg}
\end{figure}

As postulated by \citet{bayer2013fast}, we observe too that the largest eigenvalue, when training with stochastic noise of dropout in recurrent weights, gets stuck at a lower spectral radius after a fixed number of epochs over multiple tries. There is less incentive for weight matrices with lower spectral radii to change their values by a bigger amount, due to the lack of error information that can be stored over longer time delays. This can be seen in Fig.~\ref{fig:spectralradius_multseq} and Fig.~\ref{fig:spectralradius_dropout}. However, this is not the case with norm-based regularizers where the spectral radius continues to grow, albeit very slowly (Fig.~\ref{fig:spectralradius_reg}). 


\begin{figure}
	\includegraphics[width=\linewidth]{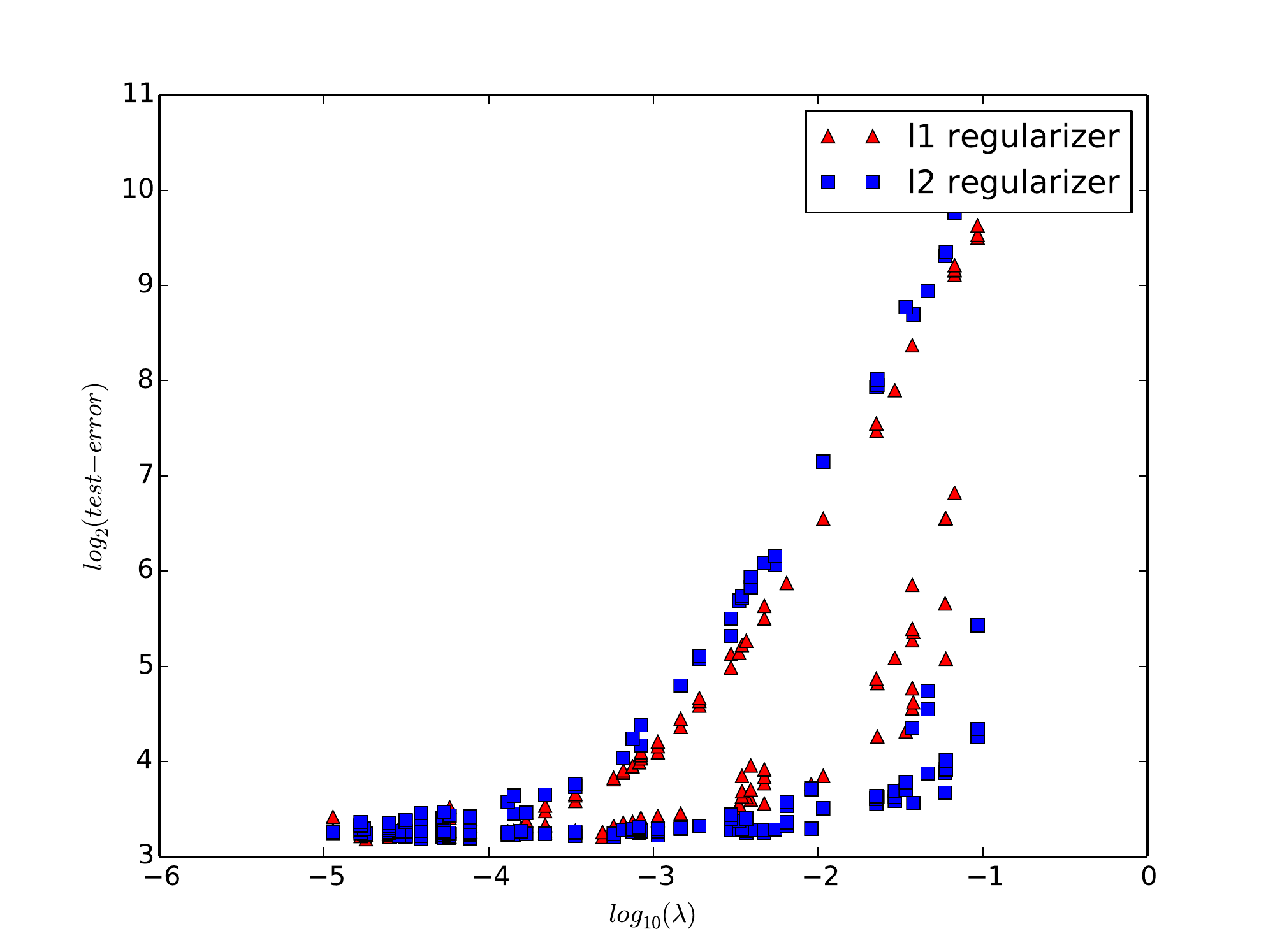}
	\caption{Regularizer $\lambda$ vs.~mean test-error for JSB Chorales}
	\label{fig:lambda_testerror}
\end{figure}

\begin{figure}
	\includegraphics[width=\linewidth]{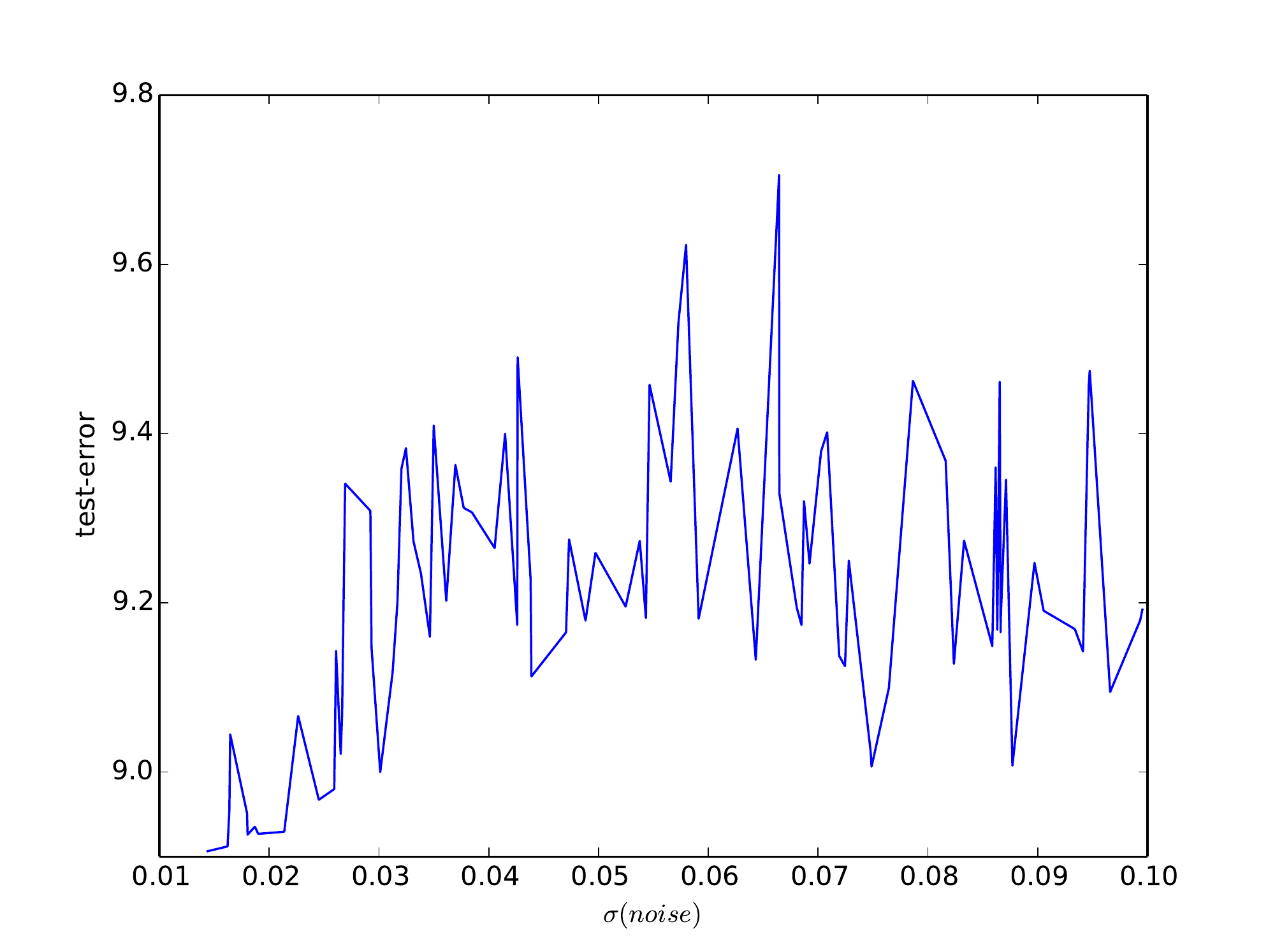}
	\caption{Additive noise $\sigma$ vs.~mean test-error for JSB Chorales}
	\label{fig:noise_testerror}
\end{figure}

\begin{figure}
	\includegraphics[width=\linewidth]{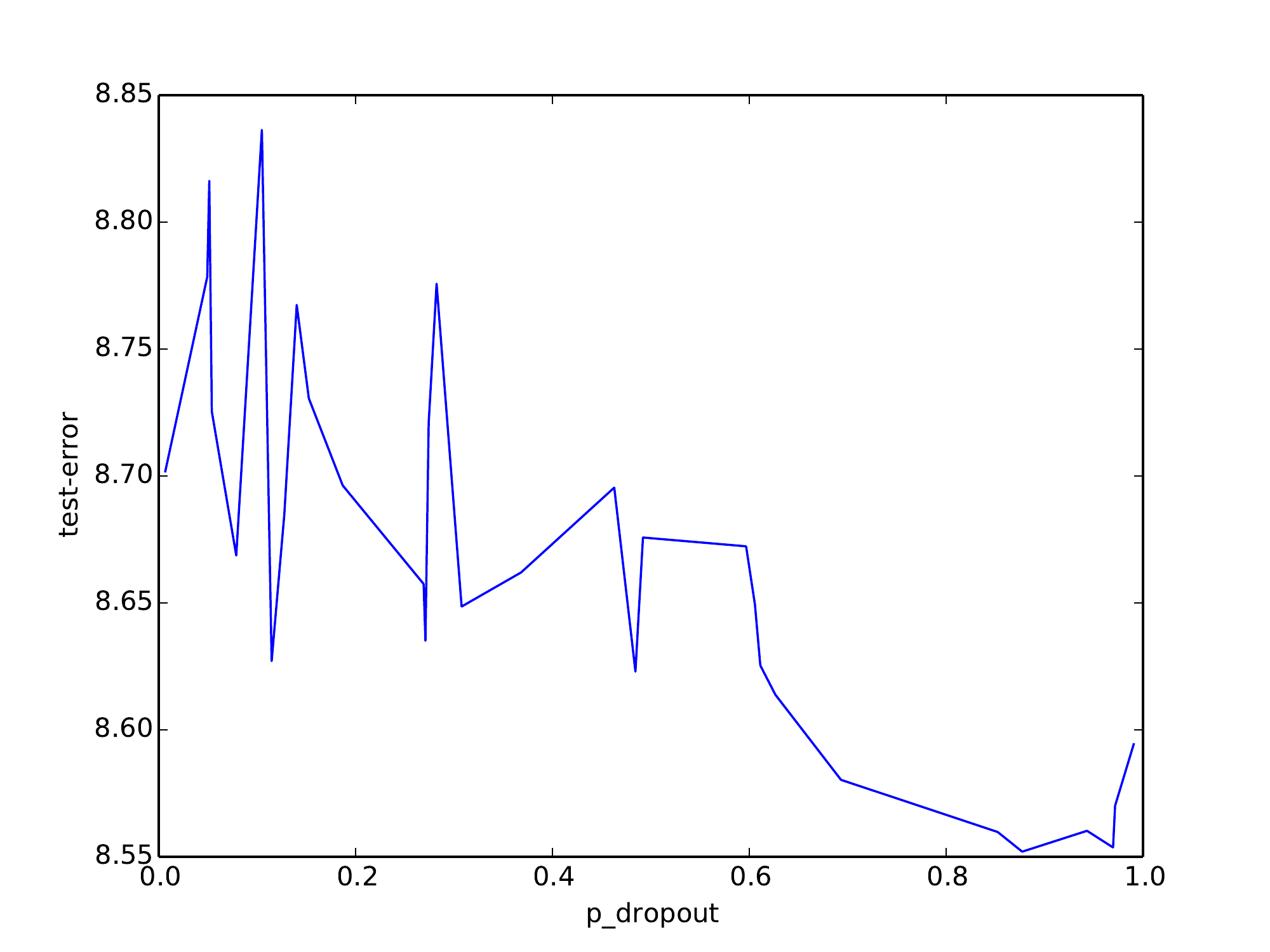}
	\caption{Dropout probability (per-sequence) vs.~mean test-error for JSB Chorales}
	\label{fig:dropout_testerror}
\end{figure}

Tab.~\ref{tab:params_random_search} in appendix \ref{Hyper params} gives the range of values from which we generate $\lambda$ for norm-based regularizer. Fig.~\ref{fig:lambda_testerror} shows the average logarithmic test errors over different $\lambda$ for both, L1 and L2, regularizers. 
Fig.~\ref{fig:noise_testerror} shows the average test errors over different $\sigma$ (standard deviation) of additive stochastic noise. The general trend indicates that the network performance decreases as $\sigma$ increases. 
Fig.~\ref{fig:dropout_testerror} shows the average test errors over different $dropout\_p$ (probability that an incoming recurrent weight is set to zero) values, for uniform dropout per-sequence. The general trend indicates that the network performance improves as $dropout\_p$ is increased. 

\section{Conclusion}
\label{Conclusion}
Through an exhaustive set of experiments with noisy weight updates, random dropout and norm-based regularization approach we have shown that conjectures about the inefficacy of MLP specific regularizers on RNNs are verifiable. \citet{pascanu2012difficulty} conjectured that a norm-based penalty on the loss function may reduce the training regime of an RNN to a single point attractor, since the length of the eigenvectors of the weight matrix never exceeded by more than a limited amount. A matrix of weights with such low spectral radius would not suffer from exploding or vanishing gradients at the cost of storing long term dependency effects. We can see this from the demonstration of a simple RNN (Fig.~\ref{fig:desirable_opt}). In fact, the analytic presentation of the noisy weight training method shows that noise in weights can also be explained as a loss regularization term. 

As the results of stochastic noise, L1 and L2 regularizers on RNNs have not been sufficiently tackled by past works in the field, we believe that we have closed a much needed empirical gap by showing that second order optimization methods, structural solutions or more sophisticated methods of training are indeed imperative to deal with the issues of vanishing gradients and long term dependency in recurrent networks.

\bibliography{report}
\bibliographystyle{icml2010}

\appendix
\appendixpage
\section{Analysis of Noisy Weights}
\label{noise_analysis}
In this section we attempt to show that adding stochastic noise to the weight matrix is equivalent to adding a regularization term to the loss function of the RNN. 

Let us define the pre-synaptic activation of the incoming connections to one hidden unit as as $a = \mathbf{w}^T\mathbf{x}$. Then, upon adding multiplicative noise to the weight vector, we have

\begin{align*}
a = (\mathbf{w} + \mathbf{\Delta_m} \mathbf{w})^T\mathbf{x}
\end{align*}

\begin{align}
E[a] &= E[(\mathbf{w} + \mathbf{\Delta_m} \mathbf{w})^T\mathbf{x}] \nonumber \\
\end{align}
The noise, $\mathbf{\Delta_m}$ is drawn from a zero mean Gaussian ($\mathbf{\Delta_m} \sim \mathcal{N}(0, \sigma)$). Additionally, considering $\mathbf{w}$ and $\mathbf{x}$ as constants, we have --
\begin{align}
\label{eqn:expecation a}
E[a] = \mathbf{w^T x}
\end{align}

This shows that the expected value of $a$ is the same as the expected value of pre--synaptic signal that is not perturbed by noise.

For computing the variance of $a$, we know that,
\begin{align*}
V[AB] = V[A]E[B]^2 + V[B]E[A]^2 + V[A]V[B]
\end{align*}

Therefore, 
\begin{align}
\label{eqn:Variance a}
V[a] &= V[(\mathbf{w}+\mathbf{\Delta_m w})^T \mathbf{x}] \nonumber \\
&= V[\mathbf{w} + \mathbf{\Delta_m w}] \mathbf{x}^2 \nonumber \\
&+ V[\mathbf{x}] E[\mathbf{w} + \mathbf{\Delta_m w}]^2  \nonumber \\
&+ V[\mathbf{w} + \mathbf{\Delta_m w}] V[\mathbf{x}]
\end{align}

The second and third terms on the right hand side of Eq.~\ref{eqn:Variance a} are zero since the variance in question is that of a constant input, $\mathbf{x}$. 

For the first term of Eq.~\ref{eqn:Variance a}, 
\begin{align}
V[\mathbf{w} + \mathbf{\Delta_m w}] = \mathbf{w}^2 \sigma^2
\end{align}

Where $\sigma$ is the standard deviation of the Gaussian noise matrix, $\Delta_m$.

Putting this back into Eq.~\ref{eqn:Variance a}, we get
\begin{align}
V[a] = \sigma^2 (\mathbf{w^T x})^2
\end{align}

The forms of $E[a]$ and $V[a]$ imply that,
\begin{align}
\Delta_m \sim \mathcal{N}(0, \sigma)
\implies a \sim \mathcal{N}(E[a], V[a])
\end{align}

This means that if the multiplicative noise is assumed to have been sampled from a Gaussian distribution, it is equivalent to assume that the pre-synaptic activations are sampled from a Gaussian. 

This equivalence to a sampling form brings us to the sampling form of pre-synaptic activation explained by \citet{bayer2013fast}, instead of smooth Gaussian approximation. 

In place of $a$, let us use $\hat{a}$, which we define as -- 
\begin{align*}
\hat{a} = E[a] + s\sqrt{V[a]}
\end{align*}
Where, $s \sim \mathcal{N}(0,1)$. 

Using the above incarnation of $a$ to it's sampling form, $\hat{a}$, we may define an effective loss function as follows --
\begin{align}
\label{eqn:effective loss}
\frac{\partial \mathcal{J}}{\partial \hat{a}}\frac{\partial \hat{a}}{\partial w_i} = \frac{\partial \mathcal{J}}{\partial \hat{a}}\left[\frac{\partial \hat{a}}{\partial E[a]}\frac{\partial E[a]}{\partial w_i} + \frac{\partial \hat{a}}{\partial V[a]}\frac{\partial V[a]}{\partial w_i}\right]
\end{align}
We will analyse the right hand side of Eq.~\ref{eqn:effective loss} one at a time. 

Consider the first term -- 
\begin{align}
\label{eqn:loss term}
\frac{\partial \mathcal{J}}{\partial \hat{a}}\frac{\partial \hat{a}}{\partial E[a]}\frac{\partial E[a]}{\partial w_i} = \frac{\partial \mathcal{J}}{\partial \hat{a}}x_i ,
\end{align}
Using the expectation value from Eq.~\ref{eqn:expecation a}. 

For a pre-synaptic activation, $\hat{a}$, Eq.~\ref{eqn:loss term} is similar to the usual backpropagation term w.r.t a loss function, $\varepsilon^{\hat{a}}$. Therefore, we may simply use the following form of the gradient term --
\begin{align}
\label{eqn:standard gradient}
\frac{\partial \varepsilon^{\hat{a}}}{\partial w_i} = \frac{\partial \mathcal{J}}{\partial \hat{a}}x_i 
\end{align}

Consider now the second term of Eq.~\ref{eqn:effective loss} --
\begin{align}
\label{eqn: variance term}
\frac{\partial \mathcal{J}}{\partial \hat{a}}\frac{\hat{a}}{\partial V[a]}\frac{\partial V[a]}{\partial w_i} &= \frac{\partial \mathcal{J}}{\partial \hat{a}}\frac{\partial \hat{a}}{\partial \sqrt{V[a]}}\frac{\partial \sqrt{V[a]}}{\partial V[a]}\frac{\partial V[a]}{\partial w_i} \nonumber \\
&= \frac{\partial \mathcal{J}}{\partial \hat{a}}\cdot s \cdot \frac{1}{2\sqrt{V[a]}}\cdot \frac{\partial (\sigma\mathbf{w^T x})^2}{\partial (\sigma\mathbf{w^T x})}\cdot \frac{\partial (\sigma\mathbf{w^T x})}{\partial w_i} \nonumber \\
&= s \sigma x_i \frac{\partial \mathcal{J}}{\partial \hat{a}}
\end{align}
This is the same as the post-synaptic gradient term, scaled by the standard deviation of the noise, $\sigma$, and independent of the actual weight values.

Hence, we can write Eq.~\ref{eqn:effective loss} as --
\begin{align}
\frac{\partial \mathcal{J}}{\partial w_i} = \frac{\partial \varepsilon^{\hat{a}}}{\partial w_i} + \frac{\partial \mathcal{R}^a_{sampling}}{\partial w_i}
\end{align}
Where the second term on the right hand side is the regularization term due to multiplicative noise addition to the synaptic weights. 

Similar analysis can be done for dropout in recurrent weight matrix, where the Gaussian distribution of the noise vector can be replaced by a Bernoulli distribution approximation when choosing $dropout\_p$. 
\section{Hyper Parameters for RNN Models}
\label{Hyper params}

For each of the eight RNN models for which the results are listed in Tab.~\ref{tab:Results-table} we generate 50 experiments with model hyper parameters chosen from the ranges given in Tab.~\ref{tab:params_random_search}.

The best configurations for all datasets are listed in Tab.~\ref{tab:JSBChorales best params}, Tab.~\ref{tab:Nottingham best params}, Tab.~\ref{tab:Piano-midi best params} and Tab.~\ref{tab:MuseData best params}.

\begin{table*}[h]
	\caption{RNN hyper parameter ranges}
	\label{tab:params_random_search}
	\begin{tabularx}{\textwidth}{|l|l|X|}
		\hline
		\multirow{4}{*}{Initilization parameters}       & $\sigma$ for $W_{hh}$                 & \{1e-3, 1, 1e-4\}    \\ \cline{2-3} 
		& $\sigma$ for $W_{ih}$ & \{1e-1, 1e-2, 1e-3\} \\ \cline{2-3} 
		& Sparsify                                     & \{15, 25, 50\}       \\ \cline{2-3} 
		& $\rho$ limit                        & \{0.9, 1.0, 1.1\}    \\ \hline
		\multirow{2}{*}{Regularizer}                    & Regularizer                                  & \{L1, L2\}           \\ \cline{2-3} 
		& Regularizer $\lambda$                          & {[}10e-2, 10e-4{]}   \\ \hline
		Dropout                                         & $dropout\_p$                     & {[}0.0, 1.0{]}       \\ \hline
		Additive and multiplicative noise               & $\sigma$ for $\Delta$              & {[}0.01, 0.1{]}      \\ \hline
		\multirow{3}{*}{Optimizer (\emph{rmsprop}) parameters} & Momentum                                     & \{0.9, 0.95, 0.99\}  \\ \cline{2-3} 
		& Step rate                                    & \{1e-2, 1e-3, 1e-4\} \\ \cline{2-3} 
		& Batch size                                   & \{27, 81\}           \\ \hline
	\end{tabularx}
	
\end{table*}

\begin{table*}[h]	
	\caption{Best configurations for JSB Chorales}
	\label{tab:JSBChorales best params}
	\begin{tabularx}{\textwidth}{|l|l|X|X|X|X|X|X|X|X|}
		\hline
		&                                              & RNN-NBR & RNN-N  & RNN-NS & RNN-MN & RNN-MNS & RNN-DO & RNN-DOS & RNN-FF \\ \hline
		\multicolumn{1}{|l|}{\multirow{4}{*}{Initilization}}       & $\sigma$ for $W_{hh}$                 & 0.0001  & 0.001  & 0.001  & 0.0001 & 0.0001  & 0.001  & 0.001   & 0.001  \\ \cline{2-10} 
		\multicolumn{1}{|l|}{}                                                & $\sigma$ for $W_{ih}$  & 0.1     & 0.1    & 0.001  & 0.01   & 0.001   & 0.01   & 0.001   & 0.1    \\ \cline{2-10} 
		\multicolumn{1}{|l|}{}                                                & Sparsify                                     & 15      & 50     & 50     & 50     & 25      & 25     & 50      & 15     \\ \cline{2-10} 
		\multicolumn{1}{|l|}{}                                                & $\rho$ limit                        & 1.1     & 0.9    & 1.0      & 0.9    & 0.9     & 1.0      & 1.0       & 0.9    \\ \hline
		\multicolumn{1}{|l|}{\multirow{2}{*}{Regularizer}}                    & Regularizer                                  & L2      & --      & --      & --      & --       & --      & --       & --      \\ \cline{2-10} 
		\multicolumn{1}{|l|}{}                                                & $\log(\lambda)$                                 & -3.93   & --      & --      & --      & --       & --      & --       & --      \\ \hline
		\multicolumn{1}{|l|}{Dropout}                                         & $dropout\_p$                     & --       & --      & --      & --      & --       & 0.92   & 0.56    & -      \\ \hline
		\multicolumn{1}{|l|}{Noise}               & $\sigma$ for $\Delta$              & --      & 0.01   & 0.04   & 0.06   & 0.01    & --      & --       & 0.09   \\ \hline
	\multicolumn{1}{|l|}{\multirow{3}{*}{Optimizer}} & Momentum                                     & 0.9     & 0.99   & 0.90    & 0.95   & 0.90     & 0.95   & 0.95    & 0.90    \\ \cline{2-10} 
	\multicolumn{1}{|l|}{}                                                & Step rate                                    & 0.001   & 0.0001 & 0.001  & 0.0001 & 0.0001  & 0.0001 & 0.0001  & 0.0001 \\ \cline{2-10} 
	\multicolumn{1}{|l|}{}                                                & Batch size                                   & 81      & 27     & 27     & 81     & 27      & 81     & 81      & 81     \\ \hline
	\multicolumn{1}{|l|}{Hidden layer}                                    & \# hidden                              & 200     & 200    & 200    & 200    & 200     & 200    & 200     & 200    \\ \hline
\end{tabularx}
\end{table*}

\begin{table*}[h]	
	\caption{Best configurations for Nottingham}
	\label{tab:Nottingham best params}
	\begin{tabularx}{\textwidth}{|l|l|X|X|X|X|X|X|X|X|}
		\hline
		&                                              & RNN-NBR & RNN-N  & RNN-NS & RNN-MN & RNN-MNS & RNN-DO & RNN-DOS & RNN-FF \\ \hline
		\multicolumn{1}{|l|}{\multirow{4}{*}{Initilization}}       & $\sigma$ for $W_{hh}$                 & 0.0001  & 0.0001  & 0.001  & 0.0001 & 0.0001  & 0.001  & 0.001   & 0.0001  \\ \cline{2-10} 
		\multicolumn{1}{|l|}{}                                                & $\sigma$ for $W_{ih}$  & 0.1     & 0.1    & 0.01  & 0.001   & 0.001   & 0.01   & 0.1   & 0.001    \\ \cline{2-10} 
		\multicolumn{1}{|l|}{}                                                & Sparsify                                     & 15      & 25     & 25     & 15     & 25      & 15     & 25      & 15     \\ \cline{2-10} 
		\multicolumn{1}{|l|}{}                                                & $\rho$ limit                        & 0.9     & 1.1    & 1.0      & 1.0    & 1.0     & 0.9      & 1.1       & 1.1    \\ \hline
		\multicolumn{1}{|l|}{\multirow{2}{*}{Regularizer}}                    & Regularizer                                  & L2      & --      & --      & --      & --       & --      & --       & --      \\ \cline{2-10} 
		\multicolumn{1}{|l|}{}                                                & $\log(\lambda)$                                 & -3.77   & --      & --      & --      & --       & --      & --       & --      \\ \hline
		\multicolumn{1}{|l|}{Dropout}                                         & $dropout\_p$                     & --       & --      & --      & --      & --       & 0.36   & 0.78    & --      \\ \hline
		\multicolumn{1}{|l|}{Noise}               & $\sigma$ for $\Delta$              & --       & 0.01   & 0.02   & 0.02   & 0.06    & --      & --       & 0.05   \\ \hline
		\multicolumn{1}{|l|}{\multirow{3}{*}{Optimizer}} & Momentum                                     & 0.95     & 0.95   & 0.95    & 0.95   & 0.95     & 0.90   & 0.90    & 0.95    \\ \cline{2-10} 
		\multicolumn{1}{|l|}{}                                                & Step rate                                    & 0.0001   & 0.0001 & 0.0001  & 0.0001 & 0.0001  & 0.001 & 0.001  & 0.0001 \\ \cline{2-10} 
		\multicolumn{1}{|l|}{}                                                & Batch size                                   & 81      & 27     & 27     & 27     & 27      & 81     & 81      & 27     \\ \hline
		\multicolumn{1}{|l|}{Hidden layer}                                    & \# hidden                              & 200     & 200    & 200    & 200    & 200     & 200    & 200     & 200    \\ \hline
	\end{tabularx}
\end{table*}

\begin{table*}[h]

	\caption{Best configurations for Piano-midi.de}
	\label{tab:Piano-midi best params}
	\begin{tabularx}{\textwidth}{|l|l|X|X|X|X|X|X|X|X|}
	\hline
	&                    & RNN-NBR & RNN-N  & RNN-NS & RNN-MN & RNN-MNS & RNN-DO & RNN-DOS & RNN-FF \\ \hline
	\multirow{4}{*}{Initialization} & $\sigma$ for $W_{hh}$ & 0.0001  & 0.001  & 0.001  & 0.0001 & 0.0001  & 0.001  & 0.0001  & 0.0001 \\ \cline{2-10} 
	& $\sigma$ for $W_{ih}$ & 0.001   & 0.1    & 0.001  & 0.001  & 0.1     & 0.001  & 0.01    & 0.1    \\ \cline{2-10} 
	& Sparsify              & 15      & 25     & 15     & 15     & 15      & 15     & 25      & 50     \\ \cline{2-10} 
	& $\rho$ limit          & 0.90     & 1.0      & 1.0      & 1.0      & 0.90     & 1.0      & 0.90     & 0.90    \\ \hline
	\multirow{2}{*}{Regularizer}    & Regularizer           & L2      & --      & --      & --      & --       & --      & --       & --      \\ \cline{2-10} 
	& $\log(\lambda)$       & -3.52   & --      & --      & --      & --       & --      & --       & --      \\ \hline
	Dropout                         & $dropout\_p$           & --       & --      & --      & --      & --       & 0.69   & 0.51    & --      \\ \hline
	Noise                           & $\sigma$ for $\Delta$ & --       & 0.05   & 0.04   & 0.04   & 0.02    & --      & --       & 0.08   \\ \hline
	\multirow{3}{*}{Optimizer}      & Momentum              & 0.95    & 0.95   & 0.99   & 0.90    & 0.95    & 0.90    & 0.95    & 0.90    \\ \cline{2-10} 
	& Step rate             & 0.0001  & 0.0001 & 0.0001 & 0.001  & 0.0001  & 0.0001 & 0.0001  & 0.0001 \\ \cline{2-10} 
	& Batch size            & 27      & 27     & 81     & 81     & 81      & 27     & 81      & 81     \\ \hline
	Hidden layer                    & \# hidden             & 100     & 100    & 100    & 100    & 100     & 100    & 100     & 100    \\ \hline
	\end{tabularx}
\end{table*}

\begin{table*}[h]
	\caption{Best configurations for MuseData}
	\label{tab:MuseData best params}
	\begin{tabularx}{\textwidth}{|l|l|X|X|X|X|X|X|X|X|}
		\hline
		&                       & RNN-NBR & RNN-N  & RNN-NS & RNN-MN & RNN-MNS & RNN-DO & RNN-DOS & RNN-FF \\ \hline
		\multirow{4}{*}{Initialization} & $\sigma$ for $W_{hh}$ & 0.001   & 0.0001 & 0.0001 & 0.001  & 0.001   & 0.0001 & 0.0001  & 0.0001 \\ \cline{2-10} 
		& $\sigma$ for $W_{ih}$ & 0.01    & 0.01   & 0.1    & 0.1    & 0.1     & 0.001  & 0.1     & 0.001  \\ \cline{2-10} 
		& Sparsify              & 25      & 50     & 15     & 50     & 50      & 50     & 25      & 15     \\ \cline{2-10} 
		& $\rho$ limit          & 1.0       & 0.9    & 1.1    & 1.0      & 1.0       & 1.1    & 1.0       & 0.9    \\ \hline
		\multirow{2}{*}{Regularizer}    & Regularizer           & L1      & --      & --      & --      & --       & --      & --       & --      \\ \cline{2-10} 
		& $\log(\lambda)$       & -3.80    & --      & --      & --      & --       & --      & --       & --      \\ \hline
		Dropout                         & $dropout\_p$           & --       & --      & --      & --      & --       & 0.93   & 0.80     & --      \\ \hline
		Noise                           & $\sigma$ for $\Delta$ & --       & 0.02   & 0.02   & 0.04   & 0.09    & --      & --       & 0.01   \\ \hline
		\multirow{3}{*}{Optimizer}      & Momentum              & 0.90     & 0.99   & 0.95   & 0.95   & 0.90     & 0.90    & 0.90     & 0.95   \\ \cline{2-10} 
		& Step rate             & 0.0001  & 0.0001 & 0.0001 & 0.0001 & 0.0001  & 0.0001 & 0.0001  & 0.0001 \\ \cline{2-10} 
		& Batch size            & 81      & 27     & 27     & 81     & 81      & 27     & 81      & 81     \\ \hline
		Hidden layer                    & \# hidden             & 600     & 600    & 600    & 600    & 600     & 600    & 600     & 600    \\ \hline
	\end{tabularx}
\end{table*}

\end{document}